\documentclass{article}

\usepackage[dblblindworkshop,final]{neurips_2026}  % <- AXIOM CAMERA-READY: de-anon + workshop footer
\workshoptitle{AXIOM Foundations of Efficient Deep Learning}

\usepackage[utf8]{inputenc}
\usepackage[T1]{fontenc}
\usepackage[hidelinks]{hyperref}
\usepackage{url}
\usepackage{booktabs}
\usepackage{amsfonts}
\usepackage{amsmath}
\usepackage{amssymb}
\usepackage{nicefrac}
\usepackage{microtype}
\usepackage{xcolor}
\usepackage{graphicx}
\usepackage{tikz}
\usetikzlibrary{positioning}

\usepackage{enumitem}

\title{The Structure of Quantization Damage in LLMs:\\
       Why the Next Bit Should Be Spent Globally}

\makeatletter
\newif\ifshowcredits
\if@anonymous
  \newcommand{\authornote}{}%
  \showcreditsfalse
\else
  \newcommand{\authornote}{\thanks{Corresponding author (\texttt{jundongh@alumni.upenn.edu}). Jundong Hu led and carried out the research end to end.}}%
  \showcreditstrue
\fi
\makeatother
\author{%
  Jundong Hu\authornote \\
  PayPal AI \\
  \texttt{jundhu@paypal.com} \\
  \And
  Shekar Ramachandran \\
  PayPal AI \\
  \texttt{sheramachandran@paypal.com} \\
}

\begin{document}
% Re-assert submission line numbers (a package loaded after neurips_2026 can deactivate lineno).
% Harmless under [final]/[preprint], where lineno is not loaded and this is a no-op via the guard below.
\ifdefined\linenumbers\linenumbers\fi
\maketitle

% =============================================================================
\begin{abstract}
Post-training quantization (PTQ) is widely used to reduce the cost of serving large language models (LLMs),
but its accuracy cost is
uneven and is often tuned per model. We study where quantization damage occurs and how to allocate a
small additional precision budget. Using causal mixed-precision intervention as
ground truth (raise each layer to 8-bit in turn and measure the accuracy it recovers) across 9 open-weight models
in 4 architecture families, we test 3 intuitive hypotheses: that quantization damage lives in task circuits, where the model computes, or in weight statistics. None of them predicts which layers benefit from restored
precision. Recovery is instead diffuse: for 8 of 9 models, recovering $75\%$ of the gap takes roughly half
the layers; the lone exception, Qwen3-8B, is sharply concentrated. At a matched precision budget, spending it globally on finer quantization granularity beats
locally repairing the most recoverable layers for all 8 group-128-compatible models (all but OpenLLaMA, whose width rules out group-128), by $21$--$52$
points, including the concentrated Qwen3-8B. We report 2 secondary findings: the residual is
budget-limited (8-bit is near-lossless in our evaluation across RTN, GPTQ, and AWQ), and the location of peak recovery correlates with architecture within a family, though not across families. Within this budget setting, global granularity is a better default than selectively protecting critical layers. More broadly, cheap signals that correlate with quantization damage do not necessarily identify where restoring precision improves accuracy; this must be tested with causal intervention.
\end{abstract}

% =============================================================================
\section{Introduction}
% BUDGET: ~0.85 page. Ends on the scoreboard; carries the whole spine.

Quantization is now standard for cheap LLM serving, but its accuracy cost is uneven: under the same 4-bit
scheme some models and tasks lose far more than others, and practitioners mitigate this by trial and error.
A natural response is to locate the damage: if we know which parts of the network low precision hurts, we can
protect them. This leads to 2 practical questions: where does the damage occur, and how should a limited
precision budget be allocated to repair it?

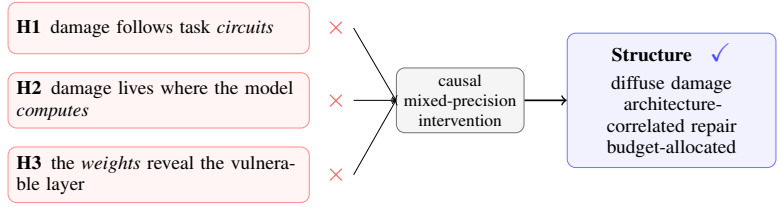
\begin{figure}[t]
  \centering
  \scalebox{0.8}{%
  \begin{tikzpicture}[
    hyp/.style={draw, rounded corners, align=left, text width=4.7cm, minimum height=0.85cm,
                font=\small, inner sep=4pt, draw=red!45, fill=red!4},
    arb/.style={draw, rounded corners, align=center, text width=1.9cm, minimum height=1.0cm,
                font=\footnotesize, inner sep=3pt, draw=black!55, fill=black!4},
    law/.style={draw, rounded corners, align=center, text width=3.1cm, minimum height=1.6cm,
                font=\small, inner sep=5pt, draw=blue!55, fill=blue!5},
    verdict/.style={font=\large\bfseries},
  ]
    \node[hyp] (h1) {\textbf{H1}\, damage follows task \emph{circuits}};
    \node[hyp, below=0.30cm of h1] (h2) {\textbf{H2}\, damage lives where the model \emph{computes}};
    \node[hyp, below=0.30cm of h2] (h3) {\textbf{H3}\, the \emph{weights} reveal the vulnerable layer};
    \node[verdict, red!70, right=0.15cm of h1] (x1) {$\times$};
    \node[verdict, red!70, right=0.15cm of h2] (x2) {$\times$};
    \node[verdict, red!70, right=0.15cm of h3] (x3) {$\times$};
    \node[arb, right=0.7cm of x2] (int) {causal\\ mixed-precision\\ intervention};
    \node[law, right=0.7cm of int] (law)
      {\textbf{Structure} \; {\large\bfseries\color{blue!70}$\checkmark$}\\[3pt]
       diffuse damage\\ architecture-correlated repair\\ budget-allocated};
    \draw[->] (x1.east) -- (int.west);
    \draw[->] (x2.east) -- (int.west);
    \draw[->] (x3.east) -- (int.west);
    \draw[->, thick] (int.east) -- (law.west);
  \end{tikzpicture}}
  \caption{\textbf{Three hypotheses, one causal test.} Three hypotheses for where quantization damage lives
    (H1--H3, left) are each refuted by mixed-precision intervention (center, $\times$); the resulting
    pattern is shown at right ($\checkmark$).}
  \label{fig:overview}
\end{figure}

We test 3 hypotheses about the location of the damage, and therefore about where precision should be
restored; 2 come from interpretability and 1 from the weights: task circuits (H1), where the model
computes (H2), or its weights (H3). Each comes with a cheap
localizer (circuit drift, causal activation patching, weight statistics), but a localizer is only useful if
it points at precision that pays off. We test all 3 against a causal ground truth,
\textbf{mixed-precision intervention}: raise each candidate location to 8-bit in turn, leave the rest at 4-bit, and
measure the accuracy recovered (its \emph{marginal value of precision}; Figure~\ref{fig:overview}).
\pagebreak
\paragraph{Contributions.}
\begin{itemize}
  \item \textbf{A resource-allocation rule for precision.} At matched effective bits/weight, global
    granularity beats oracle-selected local layer repair for all 8 group-128-compatible models
    (by $21$--$52$ points), including the most concentrated one (Qwen3-8B): unconditional
    within the tested budget setting (\S\ref{sec:disc3}).
  \item \textbf{No cheap localizer predicts the marginal value of precision.} Task circuits, the causal
    computation site, and weight/reconstruction statistics all fail to locate where restoring precision
    recovers accuracy; only the causal intervention identifies where restoring precision improves accuracy
    (\S\ref{sec:disc2}).
  \item \textbf{The structure of recovery.} Damage is diffuse (${\sim}$half the layers); the residual is
    budget-limited (8-bit near-lossless across RTN, GPTQ, AWQ); and where recovery concentrates, the site is
    architecture-correlated within a family (leave-one-out predicts the held-out LLaMA-3.x size, 3/3), not
    across (\S\ref{sec:disc2}--\S\ref{sec:disc3}).
\end{itemize}

% =============================================================================
\section{Related Work}
% BUDGET: ~0.2 page.

Standard 4-bit PTQ (round-to-nearest (RTN), group-wise scaling, GPTQ~\citep{frantar2022gptq} and
AWQ~\citep{lin2024awq}) and near-lossless \emph{mixed-precision} 8-bit inference~\citep{dettmers2022llmint8}
define our operating range. We introduce no quantization method, instead separating the
contributions of granularity and quantization algorithm. Sensitivity-based allocation (Hessian importance~\citep{dong2019hawq}, weight magnitude),
attention-aware layerwise mixed-precision~\citep{guan2024aptq}, and
salience-based weight protection~\citep{xiao2025taskcircuit} all assume damage is localizable. We find the
recoverable layer is dissociated from causal importance, weight statistics, and reconstruction error (we do
not test Hessian sensitivity itself; see \S\ref{sec:discussion}), so such heuristics transfer only within
some families, and our ``no few-layer fix'' concerns whole layers, leaving weight-level protection a separate
axis. Task-aware quantization ties damage to calibration data~\citep{williams2024calibration} and
task~\citep{levi2025youhadonejob}; we examine why it is task-specific. We also use circuit analysis and
causal-intervention methods~\citep{meng2022rome,wang2023ioi,vig2020causal}, which were developed to explain
behavior, to generate candidate localizers. We retain only signals that pass causal validation.

% =============================================================================
\section{Experimental Setup}\label{sec:setup}
% BUDGET: ~0.5 page.

\paragraph{Models and evaluation.} We study 9 open-weight models spanning 4 architecture families
and a $16\times$ size range: LLaMA-3.2-1B/3B~\citep{meta2024llama32modelcard} and
LLaMA-3-8B~\citep{llama3modelcard}; Qwen2.5-0.5B~\citep{qwen2024qwen25} and
Qwen3-0.6B/1.7B/8B~\citep{qwen2025qwen3technicalreport}; Mistral-7B~\citep{jiang2023mistral7b}; and
OpenLLaMA-3B~\citep{openlm2023openllama}. Each is evaluated on 22 tasks at 200 samples/task with a fixed
continuation-scoring harness (CORE~\citep{li2024dclm}, a nanochat-style loop~\citep{karpathy2025nanochat},
seed 1337). The tasks span reading comprehension, commonsense completion, factual retrieval, and formal/symbolic
reasoning, among others (e.g.\ \texttt{squad}, \texttt{hellaswag}, \texttt{arc\_challenge},
\texttt{dyck\_languages}), and cluster into roughly 5 reproducible groups stable across all 9 models, so
``quantization disrupts task circuits'' is a natural hypothesis (Appendix~\ref{app:f1}). \emph{Quantization damage} is a
model's fp16$\to$4-bit CORE gap.

\paragraph{Three hypotheses, one causal test.} The 3 hypotheses locate the damage, each with a cheap
\emph{localizer} computed on a common calibration set: \textbf{H1 (circuits)}, each
task's head-deviation profile (every head's activation relative to the cross-task mean), from which we
measure circuit \emph{drift}; \textbf{H2 (computation site)}, \emph{causal activation
patching}~\citep{meng2022rome}, i.e.\ which layers the prediction causally depends on; and \textbf{H3
(weights)}, per-layer weight statistics (standard deviation and reconstruction error). We compare these localizers with one causal instrument,
\textbf{mixed-precision intervention}: raise a chosen layer (or set) to 8-bit with the rest at 4-bit and read
the fraction of the 4-bit$\to$8-bit gap it recovers, the \emph{marginal value of precision} at that
location. We regard a localizer as successful only when it identifies a location at which the intervention shows a precision benefit.

\paragraph{Quantization.} The controlled damage probe is per-row RTN at 4-bit (localization,
\S\ref{sec:disc2}); the budget analysis (\S\ref{sec:disc3}) adds package GPTQ/AWQ
(\texttt{llm-compressor}~\citep{llmcompressor2024}, W4A16, group-128) and an 8-bit tier. Our group-128
configuration needs a width divisible by 128, which excludes OpenLLaMA (intermediate dim $8640$); the budget analysis
(\S\ref{sec:disc3}) therefore covers the 8 remaining models.

% =============================================================================
\section{Three Hypotheses, None Sufficient: The Damage Is Diffuse}\label{sec:disc2}
% BUDGET: ~1.15 page. First half of the scoreboard arc: H1/H2/H3 refuted -> diffuse -> concentrated.

\paragraph{None of the three localizers predicts layer-level recovery.} \textbf{H1 (circuits):} if quantization hurt a task by
reorganizing its circuit, tasks whose circuits \emph{drift} more should lose more accuracy. This prediction
fails: the raw drift--damage correlation ($r{=}{+}0.377$, $198$ task--model pairs) is a task-type
confound that falls to ${+}0.05$ (n.s.) after jointly controlling for model and category,
TOST (two one-sided tests)-equivalent to zero within
$\pm0.2$~\citep{lakens2017equivalence,schuirmann1987tost} and not underpowered (magnitude drift does
predict damage under the same test; Appendix~\ref{app:f2}). \textbf{H2 (computation site):} causal activation
patching places computation at the boundary layers (first and last MLP), which suggests protecting
them. However, the boundary pair recovers ${\le}13\%$ of the gap in 6 of 9 models (only Mistral-7B is
substantially helped, $55.5\%$; Qwen3-1.7B and OpenLLaMA exceed $13\%$ only modestly; Appendix~\ref{app:h2}),
so the computation site does not predict the repair site. \textbf{H3 (weights):} weight statistics (standard
deviation or reconstruction error) should identify the recoverable layer, but neither does so outside
LLaMA: a weight-std ``fragility'' heuristic finds it only within LLaMA (there, L1), missing Mistral's
endpoint and Qwen's L4, while reconstruction error is negatively related to recovery in the one concentrated model
(Qwen3-8B), whose most-recoverable layers are among the lowest-error ones (protecting its top 3 cuts
reconstruction error by only ${\sim}7\%$ yet recovers nearly all the accuracy; Appendix~\ref{app:h3}). We do
not test Hessian sensitivity (\S\ref{sec:discussion}).

\paragraph{The damage is diffuse.} Because none of the localizers identifies a small set of layers, we
next measure how many layers must be restored. Ranking layers by their own protect-one recovery (raising that layer alone to 8-bit; an
\emph{oracle} order) and restoring the top-$k$ to 8-bit, recovery accrues slowly: reaching $50/75/90\%$ of
the gap takes ${\sim}20/49/73\%$ of layers on average (9/9 models; curves in Figure~\ref{fig:carrier},
Appendix~\ref{app:diffuse}).
No single layer accounts for more than ${\sim}44\%$ of the damage in any model. Recovering most
of the gap thus needs roughly half the network, not a few layers.

\paragraph{Peak recovery locations correlate within a family.} Qwen3-8B is an exception:
its 3 highest-recovery layers recover essentially the whole gap (its single best only ${\sim}40\%$), so
concentration is family-contingent. Where it sits is architecture-correlated: LLaMA-3.x peaks at the same
layer (L1) at every size, so leave-one-out predicts the held-out size (3/3); Mistral (n=1) peaks at its last,
Qwen only at 8B. But this is within-family only: Qwen3's peak moves with scale, and no signal forecasts a new
family's site before a sweep (Appendix~\ref{app:diffuse}). The recoverable fraction therefore does not align
with any of the 3 localizers. Under our greedy, recovery-ranked protocol, no small layer set closes the
remaining gap; identifying layers is therefore insufficient, and we next examine allocation of the global bit
budget.

% =============================================================================
\section{Where the Next Bit Should Go: At a Matched Budget, Global Granularity Beats Local Repair}\label{sec:disc3}
% BUDGET: ~0.75 page. Second half of the scoreboard arc. Carries Table 1 (equal-budget).

Starting from per-row 4-bit RTN, a small increment of ${+}0.146$ effective bits/weight can be allocated in
2 ways: \emph{globally}, by using finer granularity everywhere (per-row $\to$ group-128 scales), or
\emph{locally}, by restoring the most-recoverable whole layers to 8-bit. We match the two at equal effective bits/weight
(first-order, weight-only, non-integer layer counts interpolated; App.~\ref{app:eqbudget}) and score recovery as \% of the per-row RTN4$\to$RTN8
CORE gap.

\paragraph{At a matched budget, global beats local for every model.} Global granularity yields higher recovery than the best matched local repair for
all 8 group-128-compatible models, by $21$--$52$ points (Table~\ref{tab:eqbudget}). The local arm is
oracle-selected (top layers by their own protect-one recovery, scored on the set they select on), so this is
a conservative bound: a deployable selector would do worse. Even the most concentrated model, Qwen3-8B,
favors global ($77$ vs.\ $55\%$). Finer granularity applied globally is therefore a better default than
selective layer protection, unconditional within the tested matched-budget setting (scope caveats in
\S\ref{sec:discussion}).

\begin{table}[t]
  \centering
  \caption{Global (g128) vs.\ matched local 8-bit repair at ${+}0.146$ eff.\ bits/weight (8
    group-128-compatible models; OpenLLaMA's dim $8640\nmid128$, so g128 is undefined). Recovery = \% of the
    per-row RTN4$\to$RTN8 CORE gap; ``top-1 conc.'' = single-best-layer recovery.}
  \label{tab:eqbudget}
  \begin{tabular}{lccc}
    \toprule
    Model & global (g128) & local @ matched & top-1 conc. \\
    \midrule
    Llama-3-8B    & $78.8$ & $44.7$ & $43.7$ \\
    Qwen3-8B      & $76.6$ & $54.9$ & $39.7$ \\
    Mistral-7B    & $72.9$ & $51.7$ & $50.9$ \\
    Qwen3-1.7B    & $66.9$ & $14.9$ & $14.9$ \\
    Llama-3.2-1B  & $64.9$ & $17.2$ & $29.5$ \\
    Llama-3.2-3B  & $62.9$ & $36.7$ & $36.5$ \\
    Qwen2.5-0.5B  & $41.1$ & $\phantom{0}3.4$ & $\phantom{0}3.9$ \\
    Qwen3-0.6B    & $33.4$ & $\phantom{0}4.8$ & $\phantom{0}4.7$ \\
    \bottomrule
  \end{tabular}
\end{table}

\paragraph{Why this holds, and that it is robust.} Concentration does not imply that local repair is better
at this budget: the local arm funds only ${\sim}1.3$ layers, and Qwen3-8B's single best recovers just
${\sim}40\%$, so even a model whose top 3 layers would close the gap cannot assemble enough of them
locally. Concentration is a property of the cumulative curve, not of what a tight budget can fund. The
comparison is also insensitive to substantial errors in the bit accounting. The equal-budget match is
first-order, but for the 7 diffuse
models local repair would need $5.16$--$6.33$ effective bits/weight to match global ($7$--$15\times$ the
disputed ${+}0.146$ increment), so the accounting would have to be wrong by $7$--$15\times$ to flip any of
them. Only Qwen3-8B is close (local ties global at $4.206$ bits/weight, ${+}0.05$ over g128), and its local
arm is oracle-selected. A task-bootstrap agrees: $P(\text{global}{>}\text{local}){\ge}0.95$ for $8/8$, and
the $95\%$ margin CI excludes local for $6/8$ (Appendix~\ref{app:eqbudget}).

\paragraph{Granularity, not algorithm, drives the 4-bit gain.} The gain from the global arm is primarily
attributable to granularity rather than calibration: per-row$\to$g128 RTN recovers ${+}0.095$ CORE on
average, whereas GPTQ and AWQ add only
${+}0.020$/${+}0.017$ over the same granularity. At 8-bit, the methods perform similarly within the evaluation noise: per-row RTN matches fp16
within harness noise (residual CI contains zero for $6/8$; full ladder, Table~\ref{tab:f5ladder8}), and every
4-bit lever collapses to zero (Appendix~\ref{app:f5}).

% =============================================================================
\section{Limitations}\label{sec:discussion}
% BUDGET: ~0.25 page.

\textbf{Untested alternatives:} our ``no few-layer fix'' is scoped to greedy, recovery-ranked interventions
at layer granularity; non-greedy layer sets and weight-level salience protection~\citep{xiao2025taskcircuit}
are untested, and H3 covers weight standard deviation and reconstruction error, not Hessian sensitivity.
\textbf{Maximal-damage regime:} localization uses per-row RTN as a maximal-damage probe (\S\ref{sec:disc2}),
so whether the same structure holds at the smaller gaps of GPTQ/AWQ/g128 is untested. \textbf{Scale ceiling:}
all runs used a single 40\,GB A100 (MIG) partition, which capped evaluation at ${\le}8$B parameters, so larger
scales are untested. \textbf{Oracle selectors:} our top-$k$ curves rank and score on the same eval set (only
Qwen3-8B is multi-seeded), so they are oracle, not deployable, selectors; small-model recoveries lie near the
noise floor, and Mistral and OpenLLaMA are single models (n=1). \textbf{First-order bit-accounting:} the
equal-budget comparison is per-layer and weight-only; an exact-byte, per-weight local allocation and a
multi-bit rate--distortion sweep~\citep{scalinglawsptq} are left to future work.

% =============================================================================
% ---- De-anonymized-only: Acknowledgments + Author Contributions, folded into the ack
%      environment so BOTH are excluded from the 4-page limit and auto-hidden in the
%      anonymous submission build (ack is \let to a hide-env under non-final). ----
\begin{ack}
\subsection*{Acknowledgments}
We thank Prakhar Mehrotra, Chandramouliswaran V, Avinash Karn, Anindya Moitra, Uma Kona, Angela McAtee, Linsey Pang, and Yun-Shiuan Chuang for their organizational support and coordination throughout this work. Jundong Hu additionally thanks Loga Vinayagam for the opportunity to join the team where this work began.

\subsection*{Author Contributions}
\textbf{Jundong Hu:} Led and carried out the research end to end, including conceptualization, methodology, implementation, experimental design and execution, analysis, and manuscript drafting and revision.

\textbf{Shekar Ramachandran:} Provided supervision, compute resources, and manuscript review.
\end{ack}

\bibliographystyle{plainnat}   % or the workshop-provided neurips_2026.bst
\bibliography{references}

% =============================================================================
\appendix
% Appendices are EXCLUDED from the 4-page limit. Reviewers not required to read them.
% Order follows the paper's logic: clusters -> drift (H1) -> protect-one/top-k + carrier (diffuse) ->
% weight stats (H3) -> budget ladder + granularity table -> equal-budget derivation.

\section{Task Circuit Clusters}\label{app:f1}

We probe each task's reliance on attention heads by recording, for every head, its activation relative to
the cross-task mean (a head-deviation profile), then cluster the 22 tasks by cosine similarity of these
profiles (Ward linkage). The cluster count is chosen by silhouette score, an internal criterion using no
task labels: sweeping $k=2\ldots8$, the silhouette is sharply maximal at $k{=}2$ (the formal-vs-rest cut)
and has a broad secondary plateau around $k{=}5$, the granularity we report as it is the finest split
that stays stable across all 9 models.

\paragraph{Five stable groups.} Across all 9 models the tasks recur in the 5 groups of
Table~\ref{tab:f1groups}. The strongest, most universal cut is $k{=}2$ (formal/structured vs.\ the rest),
drawn identically by every model across a $16\times$ size range. Head-only silhouette at $k{=}5$ ranges
$0.48$--$0.60$, and head-only clustering beats head$+$MLP in 8/9 models (adding MLP neurons degrades
quality, e.g.\ Llama-3-8B $0.54\!\to\!0.43$), so we use the head-deviation signal in the remaining analyses.

\begin{table}[h]
  \centering
  \caption{The 5 data-driven task groups (recurring across the 9 models; coqa co-clusters in 7/9, with 2
    further variable tasks discussed in the text).}
  \label{tab:f1groups}
  \begin{tabular}{ll}
    \toprule
    Group & Tasks \\
    \midrule
    Formal / structured       & dyck\_languages, lsat\_ar, cs\_algorithms, operators, repeat\_copy\_logic \\
    Short-context completion  & copa, openbook\_qa, lambada, winograd, winogrande \\
    Reading comprehension     & squad, boolq (coqa 7/9) \\
    Factual retrieval         & jeopardy, bigbench\_qa\_wikidata \\
    Multiple-choice selection & arc\_easy, arc\_challenge, hellaswag, piqa \\
    \bottomrule
  \end{tabular}
\end{table}

\paragraph{Interpretation of the clusters.} The groupings are more consistent with task \emph{demands} than with surface labels:
arc\_easy and arc\_challenge co-cluster in 9/9 models (a single multiple-choice mechanism);
language\_identification clusters with formal tasks in 7/9 (processed as pattern-matching); and
commonsense\_qa is a persistent orphan (7 different partner sets). ``Same tasks cluster'' means similar
\emph{relative} activation patterns, not that identical physical heads fire; each model finds its own
circuits under similar computational demands.

\section{Circuit Drift Is a Task-Type Confound (H1)}\label{app:f2}

\begin{figure}[h]
  \centering
  \includegraphics[width=\linewidth]{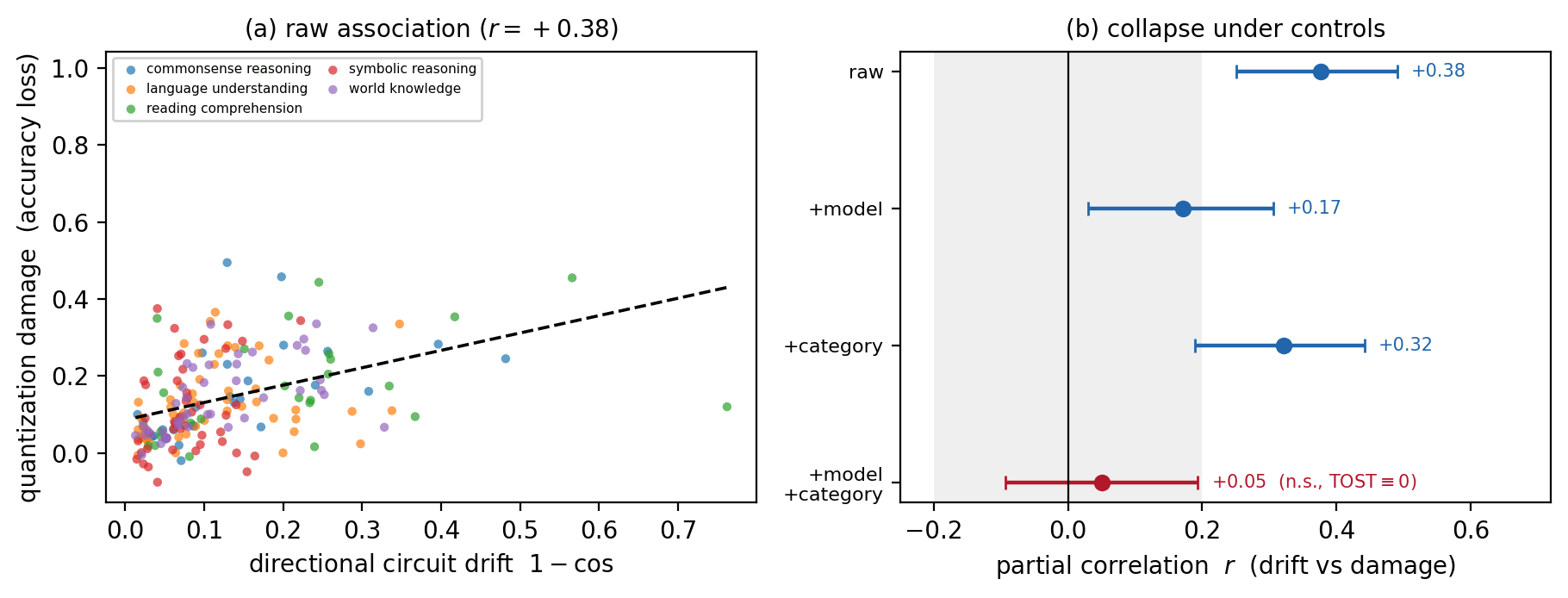}
  \caption{Circuit drift vs.\ damage. (a)~Pooled scatter of drift against damage across 198 (task,
    model) pairs, with the raw regression line. (b)~Partial correlation between drift and damage as
    controls are added left-to-right (TOST $\pm0.2$ equivalence band shaded), ending with model
    \emph{and} task category held jointly constant.}
  \label{fig:dissociation}
\end{figure}

Directional drift is $1-\cos(\mathbf{c}^{\text{fp16}},\mathbf{c}^{\text{rtn4}})$ on the centered
head-deviation profile $\mathbf{c}$.
Table~\ref{tab:f2grid} reports the full control-subset grid behind H1 (\S\ref{sec:disc2}): the drift--damage
association remains significant after each control is applied separately, but is no longer significant when
model and task category are controlled jointly (Figure~\ref{fig:dissociation}). Under joint control it is
also TOST-equivalent to zero within $\pm0.2$.
Baseline difficulty is a mild suppressor (controlling it alone \emph{raises} $r$).

\begin{table}[h]
  \centering
  \caption{Partial Pearson $r$ between directional drift and damage under increasing controls (198
    task--model pairs); each row adds the controls named at left.}
  \label{tab:f2grid}
  \begin{tabular}{lccc}
    \toprule
    Controls held constant           & partial $r$        & 95\% CI          & $p$ \\
    \midrule
    none (raw)                        & $+0.377$           & $[+0.25, +0.49]$ & $<0.001$ \\
    baseline only                     & $+0.425$           & $[+0.30, +0.53]$ & $<0.001$ \\
    category only                     & $+0.322$           & $[+0.19, +0.44]$ & $<0.001$ \\
    model only                        & $+0.172$           & $[+0.03, +0.31]$ & $0.018$  \\
    model $+$ baseline                & $+0.163$           & $[+0.02, +0.30]$ & $0.025$  \\
    \textbf{model $+$ category}       & $\mathbf{+0.054}$  & $[-0.09, +0.20]$ & $0.46$   \\
    model $+$ baseline $+$ category   & $+0.054$           & $[-0.09, +0.20]$ & $0.46$   \\
    \bottomrule
  \end{tabular}
\end{table}

A further 4 corroborating tests agree with the grid. A mixed-effects model
($\text{acc\_drop}\sim\text{drift}+(1|\text{model})$) gives a coefficient of $-0.235$ ($p=0.015$)
\emph{before} category control. The confound lives in task type. A within-model permutation test
($10^4$ permutations) gives $p=0.024$: it controls for model but not category, so it stays significant
via the task-type confound. A model-level $t$-test (each model one unit) gives mean per-model $r=+0.18$
($t=3.01$, $p=0.017$), $8/9$ positive but $0/9$ individually significant. Finally, repeating the
model-controlled test \emph{within} each of the 5 task clusters yields $|r|\le0.23$ in all 5 (all
n.s.), so the dissociation is not a quirk of one task type.

\section{H2: Where the Model Computes, Boundary-Pair Recovery}\label{app:h2}

Causal activation patching localizes computation to the boundary layers (first and last MLP); we therefore
test protection of that pair. Protecting the causal boundary pair at 8-bit recovers, per model: Mistral-7B
$55.5\%$ (L0+31); Qwen3-1.7B $21.6\%$ (L0+27); OpenLLaMA $16.3\%$ (L0+25); Qwen3-8B $12.9\%$ (L0+35);
Llama-3.2-1B $11.2\%$ (L0+15); Qwen3-0.6B $7.8\%$ (L0+27); Llama-3.2-3B $5.3\%$ (L0+27); Llama-3-8B $2.2\%$
(L0+31); Qwen2.5-0.5B $-4.1\%$ (L0+23). Overall, 6 of 9 are ${\le}13\%$ of the gap; only Mistral-7B is
substantially helped (Qwen3-1.7B and OpenLLaMA exceed $13\%$ only modestly), so the computation site does not
predict the repair site.

\section{H3: Weight Statistics Do Not Localize the Layer}\label{app:h3}

\paragraph{Weight-std vs.\ ground truth.} The weight-std ``fragility'' pair matches the protect-one peak
(Appendix~\ref{app:diffuse}) only in the LLaMA family (Table~\ref{tab:f34std}).

\begin{table}[h]
  \centering
  \caption{E1 fragile-pair recovery vs.\ the true protect-one sweep peak; ``found?'' marks whether the
    weight-std pair coincides with the sweep peak.}
  \label{tab:f34std}
  \begin{tabular}{lccc}
    \toprule
    Model & weight-std pair (recovery) & true peak & found? \\
    \midrule
    Llama-3.2-1B & L1+5 ($45.8\%$)   & L1  & yes \\
    Llama-3.2-3B & L1+5 ($37.0\%$)   & L1  & yes \\
    Llama-3-8B   & L1+15 ($40.6\%$)  & L1  & yes \\
    Mistral-7B   & L3+2 ($-1.9\%$)   & L31 & no (caught via boundary, $55.5\%$) \\
    Qwen3-8B     & L24+23 ($-0.1\%$) & L4  & no \\
    Qwen3-1.7B   & L26+21 ($6.6\%$)  & L2  & no \\
    \bottomrule
  \end{tabular}
\end{table}

\paragraph{Reconstruction-error decoupling.} Because we protect by protect-one recovery, the
recoverable layers are the \emph{lowest}-reconstruction-error layers: for Qwen3-8B, protecting the top 3
cuts total weight reconstruction error by only ${\sim}7\%$ yet recovers nearly all the accuracy, while the
remaining $285\times$ reduction buys essentially nothing. Reconstruction error increases toward later
layers, whereas CORE recovery is concentrated earlier in the network.

\section{Diffuseness and Concentration of Recovery}\label{app:diffuse}

\paragraph{No few-layer fix (per model).} Table~\ref{tab:f34single} gives the best single-layer and best
tested-pair recovery (\% of the RTN4$\to$RTN8 CORE gap, all @200/task).

\begin{table}[h]
  \centering
  \caption{Best single-layer (protect-one) and best tested-pair (E1) recovery, \% of the RTN4$\to$RTN8
    gap. Residual $= 100 - \max(\text{best single},\text{best pair})$. $^{*}$Qwen3-8B's tested pairs missed
    its sweep peak L4; we report the protect-one L4 recovery (3-seed mean $39.7{\pm}3\%$; per seed
    $38.1/44.2/36.7\%$). $^{\dagger}$open\_llama gap $0.043$ (near-noise).}
  \label{tab:f34single}
  \begin{tabular}{lccc}
    \toprule
    Model & best single (@layer) & best pair (E1) & residual \\
    \midrule
    Mistral-7B      & $50.9\%$ (L31) & $55.5\%$ (L0+31)       & ${\sim}45\%$ \\
    Llama-3-8B      & $43.7\%$ (L1)  & $40.6\%$ (L1+15)       & ${\sim}56\%$ \\
    Qwen3-8B        & $39.7{\pm}3\%$ (L4)$^{*}$  & $12.9\%$ (L0+35) & ${\sim}60\%$ \\
    Llama-3.2-3B    & $36.5\%$ (L1)  & $41.2\%$ (L0+1)        & ${\sim}59\%$ \\
    Llama-3.2-1B    & $29.5\%$ (L1)  & $45.8\%$ (L1+5)        & ${\sim}54\%$ \\
    Qwen3-1.7B      & $14.9\%$ (L2)  & $21.6\%$ (L0+27)       & ${\sim}78\%$ \\
    open\_llama\_3b & $9.2\%$ (L6)   & $22.7\%^{\dagger}$     & --- \\
    Qwen3-0.6B      & $4.7\%$ (L0)   & $7.8\%$                & ${\sim}92\%$ \\
    Qwen2.5-0.5B    & $3.9\%$ (L2)   & $4.7\%$                & ${\sim}95\%$ \\
    \bottomrule
  \end{tabular}
\end{table}

\paragraph{Damage per layer is small (single-layer damage).} No single layer accounts for more than ${\sim}44\%$
of the damage (all 9 models): max single-layer damage is Mistral $43.9\%$, Llama-3-8B $30.7\%$,
Llama-3.2-3B $28.6\%$, Llama-3.2-1B $20.5\%$, open\_llama $14.8\%$, Qwen2.5-0.5B $13.4\%$, Qwen3-8B
$13.1\%$, Qwen3-0.6B $11.5\%$, Qwen3-1.7B $8.1\%$, consistent with damage being distributed across layers.

\paragraph{E1$\leftrightarrow$E2 additivity (validation).} Pair recovery closely matches the sum of the
2 single-layer recoveries (ratios $0.91$--$1.10$): Llama-3.2-1B $45.8$ vs $44.6$ ($1.03$); Llama-3-8B
$40.6$ vs $41.0$ ($0.99$); Qwen3-1.7B $21.6$ vs $22.3$ ($0.97$); Llama-3.2-3B $41.2$ vs $37.4$ ($1.10$);
Mistral-7B $55.5$ vs $60.7$ ($0.91$). The pair results are therefore consistent with the sum of the 2 single-layer effects.

\paragraph{Top-$k$ cumulative protection.} Ranking layers by protect-one recovery and protecting the
top-$k$ together, the mean depth to recover $50/75/90\%$ of the gap is $20/49/73\%$ of layers (9/9).
Qwen3-8B is the most concentrated model: its top 3 most-recoverable layers (L4, L2, L6) recover
essentially the whole gap. Multi-seeded (eval seeds $1337/42/0$): $k1{=}39.7\pm3$, $k2{=}88.2\pm2$,
$k3{=}102.2\pm4$ (per-seed spreads are tight, suggesting that the single-layer variance is driven more by task sampling than by the evaluation seed);
the slight $>100\%$ at $k{=}3$ is eval-sampling noise around the 8-bit ceiling (the $k3$ model reconstructs
weights $285\times$ worse than full 8-bit yet scores as high; few-shot self-exclusion holds).

\begin{figure}[h]
  \centering
  \includegraphics[width=0.85\linewidth]{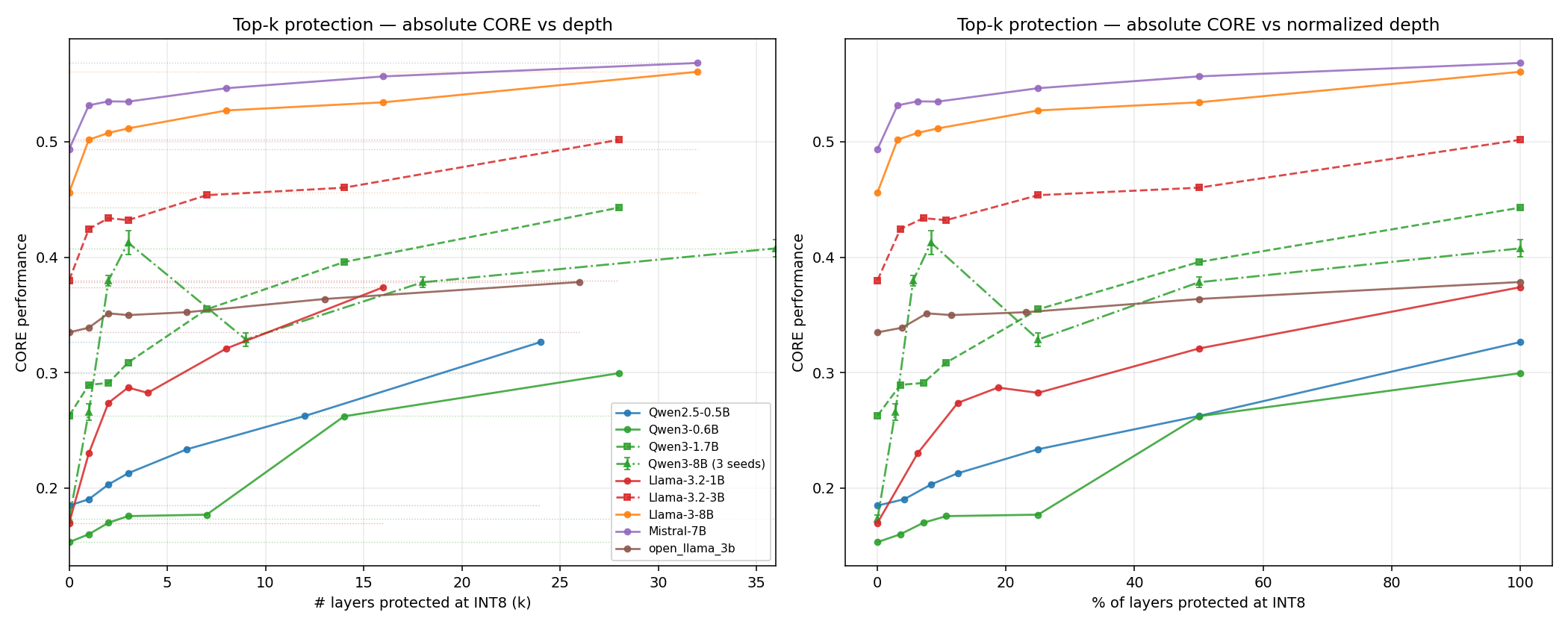}
  \caption{Cumulative CORE as the top-$k$ most-recoverable layers are restored to
    8-bit (rest at 4-bit), against each model's 4-bit (floor) and 8-bit (ceiling) bands. Layers are
    ranked by their own protect-one recovery, an \emph{oracle} ordering, not a deployable selector.
    Curves are \emph{absolute} CORE on the finite 200-sample subset, so
    cross-model levels track fp16 baselines (Qwen3-8B's level sits marginally below Qwen3-1.7B's here, a
    small-sample ordering, not a model defect); all claims are within-model recovery. Color${=}$family,
    marker/linestyle${=}$size.}
  \label{fig:carrier}
\end{figure}

\section{Budget and Method Decomposition}\label{app:f5}
\paragraph{The full ladder (CORE @200; GPTQ/AWQ = 3-seed calibration mean).} Table~\ref{tab:f5ladder}
gives the per-model rungs and the granularity/method decomposition.

\begin{table}[h]
  \centering
  \caption{Per-model 4-bit ladder. gran$\Delta$ = per-row$\to$g128 RTN (granularity); the last column is
    the GPTQ / AWQ gain over g128-RTN (method). $^{\diamond}$Qwen3-8B GPTQ exceeds fp16 (200-sample noise
    $+$ regularization); this model is excluded from \emph{both} method means for consistency.
    $^{\dagger}$open\_llama: dim $8640$ is not a
    multiple of 128, so g128 is not run; per-row RTN and fp16 only.}
  \label{tab:f5ladder}
  \begin{tabular}{lccccccc}
    \toprule
    Model & per-row & g128 & GPTQ & AWQ & fp16 & gran$\Delta$ & GPTQ / AWQ method \\
    \midrule
    Qwen2.5-0.5B          & .185 & .243 & .268 & .256 & .325 & $+.058$ & $+.025$ / $+.013$ \\
    Qwen3-0.6B            & .153 & .202 & .237 & .248 & .304 & $+.049$ & $+.035$ / $+.046$ \\
    Qwen3-1.7B            & .263 & .383 & .397 & .385 & .442 & $+.121$ & $+.013$ / $+.002$ \\
    Qwen3-8B$^{\diamond}$ & .171 & .359 & .470 & .370 & .411 & $+.187$ & $+.112$ / $+.012$ \\
    Llama-3.2-1B          & .170 & .302 & .336 & .332 & .375 & $+.133$ & $+.034$ / $+.029$ \\
    Llama-3.2-3B          & .380 & .457 & .481 & .476 & .502 & $+.077$ & $+.024$ / $+.019$ \\
    Llama-3-8B            & .456 & .538 & .539 & .548 & .556 & $+.082$ & $+.001$ / $+.009$ \\
    Mistral-7B            & .493 & .548 & .557 & .549 & .566 & $+.054$ & $+.009$ / $+.001$ \\
    open\_llama\_3b$^{\dagger}$ & .335 & --- & --- & --- & .378 & --- & --- \\
    \bottomrule
  \end{tabular}
\end{table}

\begin{table}[h]
  \centering
  \caption{Mean CORE gains over the 8 group-128-compatible models, per lever, at the 4-bit and 8-bit
    tiers. Both the GPTQ and
    AWQ means exclude Qwen3-8B (its GPTQ value exceeds fp16, a 200-sample outlier; dropped from both method
    columns for consistency); including it, GPTQ is $+0.032$ and AWQ $+0.016$.}
  \label{tab:budget}
  \begin{tabular}{lcc}
    \toprule
    Lever & 4-bit & 8-bit \\
    \midrule
    gap to fp16 @ per-row RTN            & $+0.151$ & $-0.001$ \\
    granularity (per-row $\to$ g128 RTN) & $+0.095$ & $-0.001$ \\
    GPTQ over g128 RTN                   & $+0.020$ & $+0.001$ \\
    AWQ over g128 RTN                    & $+0.017$ & $+0.000$ \\
    \bottomrule
  \end{tabular}
\end{table}

\paragraph{Headline.} Mean granularity gain $+0.095$ (Table~\ref{tab:budget}); excluding Qwen3-8B from \emph{both} method
columns (its GPTQ value exceeds fp16; Table~\ref{tab:f5ladder}), mean method gain is GPTQ $+0.020$ / AWQ
$+0.017$ ($0.21\times$ / $0.18\times$ the granularity gain); including it, GPTQ rises to $+0.032$
($0.33\times$) and AWQ is $+0.016$. Method-isolated
recovery (also excl.\ that model) is ${\approx}34\%$ (GPTQ) / ${\approx}28\%$ (AWQ) of the
fp16$\to$g128-RTN gap. Effective bits
per weight: per-row RTN $4.01$, g128 $4.156$, fp16 $16$.

\begin{figure}[h]
  \centering
  \includegraphics[width=0.9\linewidth]{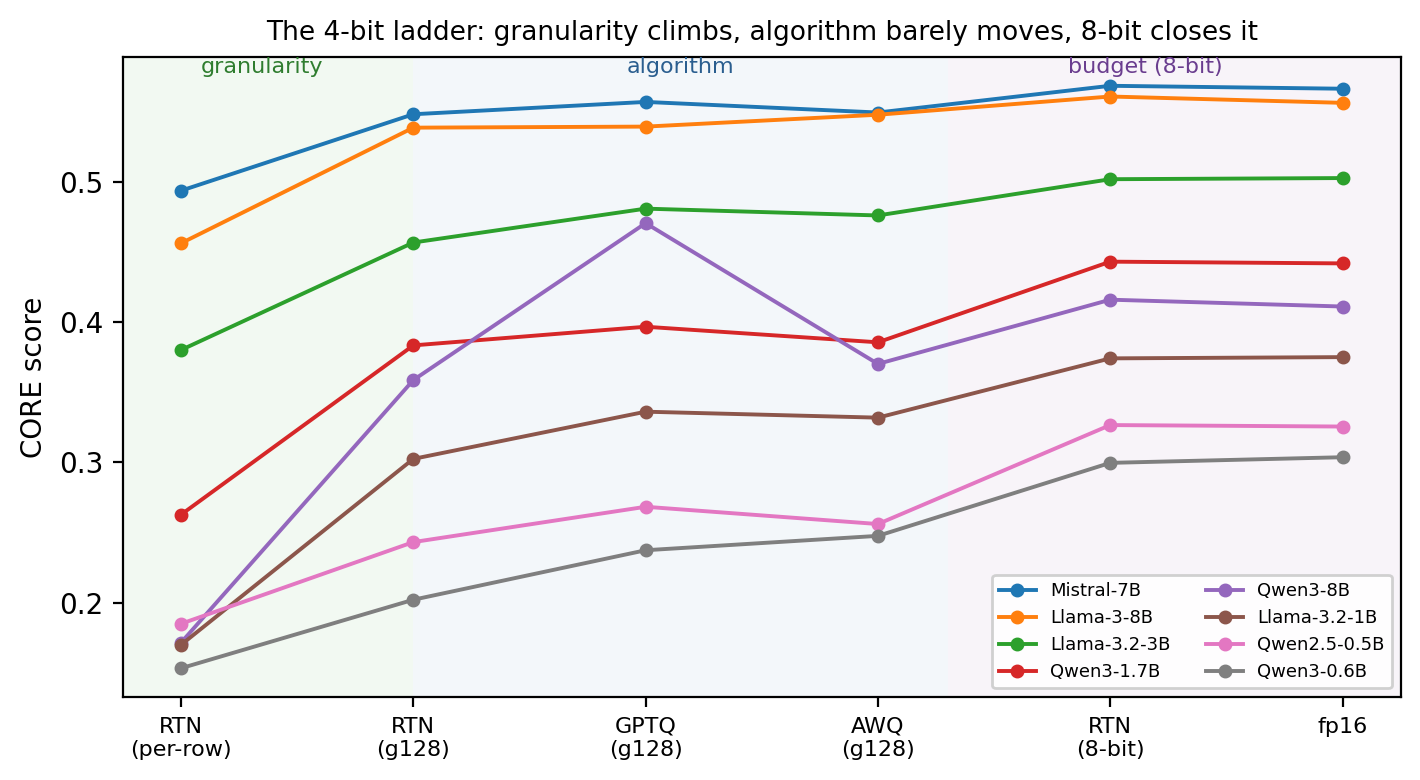}
  \caption{Per-model CORE across the ladder (per-row RTN $\to$ g128 RTN $\to$ GPTQ $\to$ AWQ $\to$
    8-bit RTN $\to$ fp16). Qwen3-8B's GPTQ spike above fp16 is the noted 200-sample outlier.}
  \label{fig:ladder}
\end{figure}

\paragraph{8-bit tier.} At 8-bit every lever collapses (Table~\ref{tab:budget}): per-row RTN
is already lossless (mean gap to fp16 $-0.001$, every model within $\pm0.005$; full per-model ladder in
Table~\ref{tab:f5ladder8}), and the Qwen3-8B GPTQ$>$fp16 anomaly resolves (GPTQ-8 $0.418\approx$ fp16
$0.411$). Figure~\ref{fig:ladder} shows the per-model ladder.

\begin{table}[h]
  \centering
  \caption{Full 8-bit ladder (CORE @200), for the 8 models with an 8-bit run. Last column: fp16 $-$
    per-row RTN8. $^{\dagger}$open\_llama was not run at 8-bit (per-row 4-bit and fp16 only;
    Table~\ref{tab:f5ladder}).}
  \label{tab:f5ladder8}
  \begin{tabular}{lcccccc}
    \toprule
    Model & fp16 & RTN8 & g128-RTN8 & GPTQ8 & AWQ8 & gap (fp16$-$RTN8) \\
    \midrule
    Qwen2.5-0.5B & .325 & .326 & .324 & .321 & .322 & $-.001$ \\
    Qwen3-0.6B   & .304 & .300 & .305 & .302 & .305 & $+.004$ \\
    Qwen3-1.7B   & .442 & .443 & .437 & .439 & .445 & $-.001$ \\
    Qwen3-8B     & .411 & .416 & .416 & .418 & .411 & $-.005$ \\
    Llama-3.2-1B & .375 & .374 & .375 & .377 & .374 & $+.001$ \\
    Llama-3.2-3B & .502 & .502 & .502 & .505 & .504 & $+.001$ \\
    Llama-3-8B   & .556 & .561 & .559 & .562 & .557 & $-.005$ \\
    Mistral-7B   & .566 & .568 & .566 & .568 & .569 & $-.002$ \\
    \bottomrule
  \end{tabular}
\end{table}

\paragraph{Seed stability and provenance.} AWQ is stable across calibration seeds (spread $0$--$2\%$);
GPTQ is more variable ($4$--$24\%$, worst on Qwen3-8B). All quantization uses llm-compressor $0.6.0.1$
(W4A16 g128), evaluated through the same CORE harness at $200$ samples/task on the fixed seed-1337
subset; GPTQ/AWQ use 3 calibration-subset seeds. open\_llama is excluded from the g128 rungs
(intermediate dim $8640 \nmid 128$) and its fp16$\to$RTN gap $0.043$ is near-noise.

\section{Equal-Budget Allocation: Global vs.\ Local (Matched Effective Bits/Weight)}\label{app:eqbudget}
At a matched increment of ${+}0.146$ effective bits/weight over per-row 4-bit RTN, we compare spending it
\emph{globally} (group-128 RTN) vs.\ \emph{locally} (restore the top layers by protect-one recovery to
8-bit, an oracle recovery ranking, rest 4-bit; matched layer count $k^{*}{=}0.0365\,n_L$, interpolated
along the top-$k$ curve). Recovery is \% of the per-row RTN4$\to$RTN8 CORE gap, so the two are directly
comparable (Table~\ref{tab:eqbudget}, main text). ``top-1'' is the single-best-layer recovery (the
concentration indicator). Global granularity yields higher recovery for all 8
models; even the most concentrated model, Qwen3-8B, favors global at this matched budget (its single most
recoverable layer recovers only ${\sim}40\%$, so the ${\sim}1.3$ layers the budget funds cannot match g128).
Note that on the shallowest model
(Llama-3.2-1B, $n_L{=}16$) the matched budget buys only $k^{*}{=}0.58$ of a layer, so its local figure
($17.2$) is the fractional-layer interpolation of its top-1 ($29.5$), i.e.\ ${+}0.146$ bits cannot even
fund one 8-bit layer there, which favors the global arm even under this interpolation.

\emph{Caveat (bounded):} the local bit-accounting is first-order (uniform-block, weight-only). To bound its effect we compute, from
each model's per-$k$ curve, the effective bits/weight local repair would need to \emph{match} global's
recovery: for the 7 diffuse models this is $5.16$--$6.33$ bits/weight, i.e.\ $1.0$--$2.2$ above g128's
$4.156$, or $7$--$15\times$ the disputed $0.146$ increment. The accounting would therefore have to be wrong by
$7$--$15\times$ to flip any of them.

Only the concentrated model, Qwen3-8B, is close: local matches global at
$4.206$ bits/weight ($+0.05$ over g128), so a ${\approx}34\%$ under-estimate of the $0.146$ increment would
tie it, but its local arm is oracle-selected (an upper bound) and its single best layer recovers only
${\sim}40\%$, so a deployable selector would need more still. Qwen3-8B's single-layer figure is stable
across eval seeds ($k{=}1$: $38.1/44.2/36.7\%$, mean $39.7{\pm}3$), with $k{=}2/3$ at $88{\pm}2$ / $102{\pm}4$
(\S\ref{app:diffuse}).

\paragraph{Task-bootstrap robustness (no new inference).} Because CORE averages over tasks and we cache
per-task scores, we resample the tasks to put confidence intervals on these margins ($5{,}000$ bootstrap
resamples of the ${\sim}22$ CORE tasks with replacement, seed $1337$). Across all 8
models, the global-favoring margin is robust: $P(\text{global}{>}\text{local})
{\ge}0.95$ for $8/8$, and the $95\%$ CI of the margin excludes the local arm for $6/8$ (the 2 exceptions,
Qwen2.5-0.5B and Mistral-7B, are the near-noise $0.5$B model and the smallest-margin singleton). We apply
the same bootstrap procedure to Qwen3-8B,
the most concentrated model. Its per-seed $k{=}1$ recovery is tight
($38.1/44.2/36.7\%$), so its uncertainty is task- rather than seed-driven, and its bootstrapped margin is
$+25$ ($95\%$ CI $[12,37]$, $P{=}1.00$). This margin is the \emph{median} of the resampled margin distribution,
which differs from the Table~\ref{tab:eqbudget} point margin of $21.7$ ($76.6{-}54.9$) because under
task-level skew the median of the per-resample margins is not the difference of the point recoveries. Applying the same resampling to the $8$-bit tier, the per-row
RTN8$\to$fp16 residual CI contains zero for $6/8$ models (8-bit is fp16-lossless within harness noise), with
a small residual ($<0.02$ CORE) remaining only for Qwen3-1.7B and Llama-3.2-1B.

\end{document}